\documentclass[lettersize,journal]{IEEEtran}
\usepackage{amsmath,amsfonts}
\usepackage{algorithmic}
\usepackage{algorithm}
\usepackage{array}
\usepackage[caption=false,font=normalsize,labelfont=sf,textfont=sf]{subfig}
\usepackage{textcomp}
\usepackage{stfloats}
\usepackage{url}
\usepackage{verbatim}
\usepackage{graphicx}
\usepackage{cite}
\usepackage{hyperref}
\hypersetup{
    colorlinks=true, % 将超链接设置为彩色
    linkcolor=red,   % 设置内部引用的超链接颜色为红色
    citecolor=blue,   % 设置引用的超链接颜色为蓝色
    urlcolor=blue    % 设置外部链接的颜色为蓝色（可选）
}

\usepackage[capitalize,noabbrev]{cleveref}
\usepackage{colortbl}
\usepackage{multirow}

\usepackage{booktabs} % for professional tables
\usepackage{xcolor} % 加载 xcolor 宏包

\begin{document}

\title{LATex: Leveraging Attribute-based Text Knowledge for Aerial-Ground Person Re-Identification}

\author{Pingping Zhang$^{*}$,\IEEEmembership{~IEEE Member}, Xiang Hu, Yuhao Wang, Huchuan Lu,\IEEEmembership{~IEEE Fellow}
        % <-this % stops a space
\thanks{Copyright (c) 2025 IEEE. Personal use of this material is permitted. However, permission to use this material for any other purposes must  be obtained from the IEEE by sending an email to \textcolor{blue}{\underline{pubs-permissions@ieee.org}}. (*Corresponding author: Pingping Zhang.)}

\thanks{X. Hu, YH. Wang and PP. Zhang are with the School of Future Technology, School of Artificial Intelligence, Dalian University of Technology, Dalian, 116024, China. (Email:1908414518@mail.dlut.edu.cn; 924973292@mail.dlut.edu.cn; zhpp@dlut.edu.cn)}

\thanks{HC. Lu is with the School of Information and Communication Engineering, Dalian University of Technology, Dalian, 116024, China. (Email: lhchuan@dlut.edu.cn)}
}

\markboth{IEEE Transactions on Intelligent Transportation Systems}%
{Shell \MakeLowercase{\textit{et al.}}: A Sample Article Using IEEEtran.cls for IEEE Journals}


\maketitle

\begin{abstract}
As an important task in intelligent transportation systems, Aerial-Ground person Re-IDentification (AG-ReID) aims to retrieve specific persons across heterogeneous cameras in different viewpoints.
Previous methods typically adopt deep learning-based models, focusing on extracting view-invariant features.
However, they usually overlook the semantic information in person attributes.
In addition, existing training strategies often rely on full fine-tuning large-scale models, which significantly increases training costs.
To address these issues, we propose a novel framework named LATex for AG-ReID, which adopts prompt-tuning strategies to leverage attribute-based text knowledge.
Specifically, with the Contrastive Language-Image Pre-training (CLIP) model, we first propose an Attribute-aware Image Encoder (AIE) to extract both global semantic features and attribute-aware features from input images.
Then, with these features, we propose a Prompted Attribute Classifier Group (PACG) to predict person attributes and obtain attribute representations.
Finally, we design a Coupled Prompt Template (CPT) to transform attribute representations and view information into structured sentences.
These sentences are processed by the text encoder of CLIP to generate more discriminative features.
As a result, our framework can fully leverage attribute-based text knowledge to improve AG-ReID performance.
Extensive experiments on three AG-ReID benchmarks demonstrate the effectiveness of our proposed methods.
The source code is available at https://github.com/kevinhu314/LATex.
\end{abstract}

\begin{IEEEkeywords}
Aerial-Ground Person Re-identification, Image-Text Retrieval, Attribute Prediction, Prompt Learning.
\end{IEEEkeywords}

\section{Introduction}
\begin{figure}[t]
    \centerline{\includegraphics[width=0.40\paperwidth]{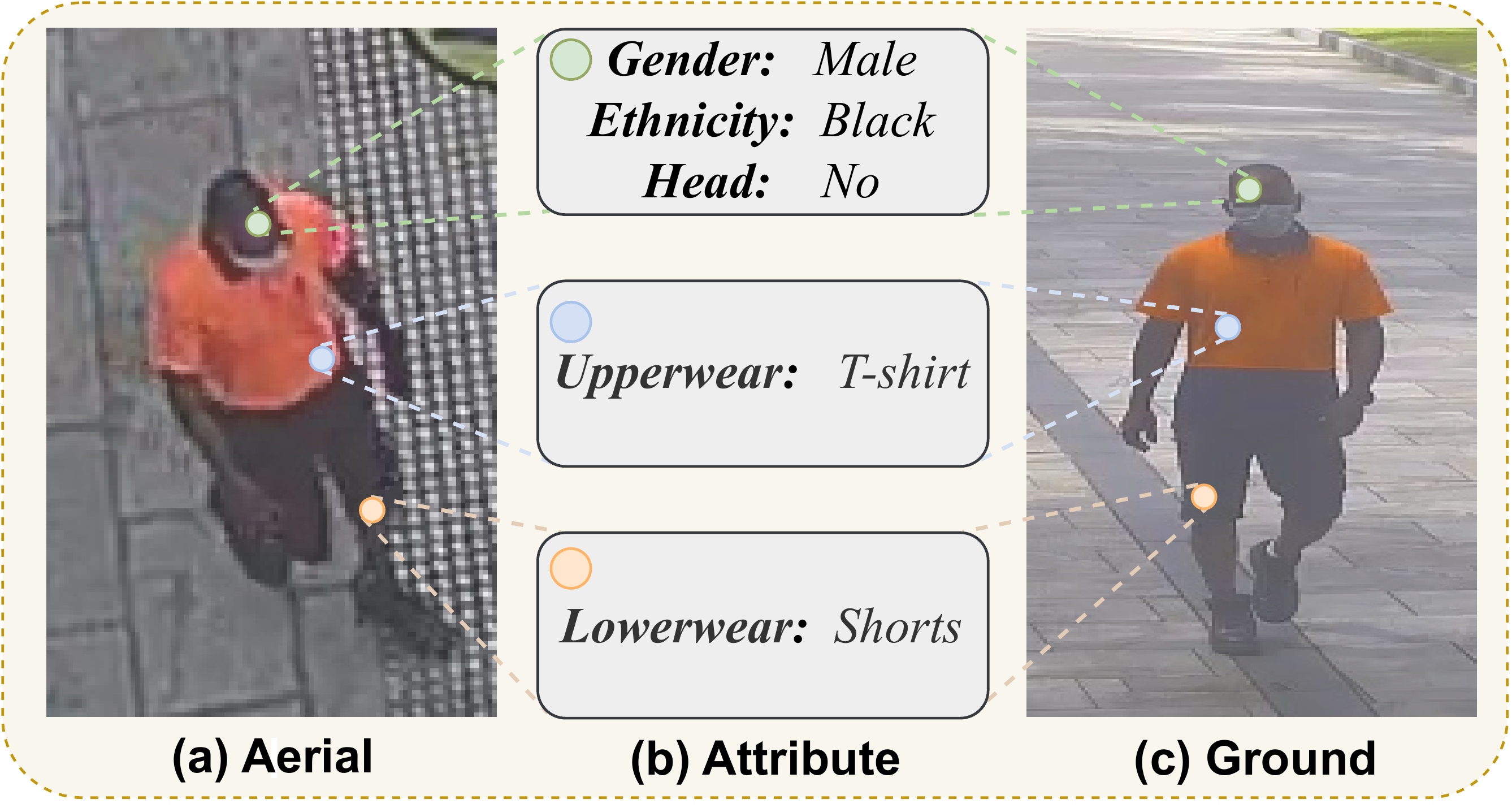}}
    \vspace{-1mm}
    \caption{An example that a person captured under (a) the aerial view by UAV and (c) the ground view by CCTV, along with (b) the corresponding person attributes. Despite significant variations in the images caused by drastic viewpoint changes, person attributes remain consistent.
    }
    \vspace{-5mm}
    \label{moti}
\end{figure}

Person Re-IDentification (ReID) aims to retrieve the same individual across different cameras.
In recent years, ReID has attracted considerable interest~\cite{lu2023learning,shi2024multi,gao2024part,liu2023video,zhang2021hat,wang2025idea} due to its wide range of applications, including intelligent surveillance and transportation system.
More recently, ReID across heterogeneous camera viewpoints, especially Aerial-Ground person ReID (AG-ReID), has become a more realistic application~\cite{nguyen2023aerial,nguyen2023ag,zhang2024view} due to the development of drones and advancements in aerial surveillance.
In practice, AG-ReID greatly helps traffic wardens to manage large transit hubs and address traffic incidents.
Unlike traditional ReID tasks~\cite{leng2019survey,li2017person}, AG-ReID amplifies the challenges posed by viewpoint variations due to drastic changes between different cameras.
These variations significantly affect the distribution of body parts, making it more difficult to learn visual features that remain consistent across diverse views.
Therefore, previous methods~\cite{nguyen2023aerial,zhang2024view} focus on mitigating the negative effects of drastic view changes and learning viewpoint-robust image features.
However, they often overlook the potential of leveraging person attributes.
As shown in Fig. \ref{moti}(a) and Fig. \ref{moti}(c), different camera viewpoints may result in significant visual differences.
Despite these significant visual differences, person attributes such as ethnicity, gender, and clothing remain unaffected.
This stability provides consistent information to obtain robust cross-view features.
Meanwhile, existing methods~\cite{nguyen2023aerial,zhang2024view} rely on the full fine-tuning strategy, significantly raising training costs.
Fortunately, prompt-tuning~\cite{jia2022visual,lester2021power} offers a way to reduce the training cost.
It also effectively integrates the pre-trained knowledge of large-scale models into specific domains~\cite{li2023clip,liu2025attribute}.

Motivated by the above observations, we propose a novel framework named LATex for AG-ReID, which leverages attribute-based text knowledge via prompt-tuning strategies to enhance the feature discrimination.
More specifically, our framework consists of three key components: an Attribute-aware Image Encoder (AIE), a Prompted Attribute Classifier Group (PACG), and a Coupled Prompt Template (CPT).
First, we introduce AIE to fine-tune the Contrastive Language-Image Pre-training (CLIP) model~\cite{radford2021learning} with learnable prompts, transferring the powerful pre-trained knowledge to AG-ReID.
In addition, AIE incorporates attribute tokens to enable fine-grained perception of person attribute information.
Then, PACG is employed to further enhance AIE's attribute perception capabilities and generate attribute representations.
Afterwards, CPT is proposed to transform attribute representations and view information into structured sentences.
Finally, these sentences are processed by CLIP's text encoder, enabling more accurate person retrieval across different camera viewpoints by explicitly leveraging information hidden in the attributes.
Extensive experiments on three AG-ReID benchmarks fully validate the effectiveness of our proposed framework.

In summary, our contributions are as follows:
\begin{itemize}
\item
\textbf{New insight.}
We observe the distinct benefits of person attributes for AG-ReID tasks.
This insight inspires us to consider the problem from an attribute-based perspective.
Based on the attribute consistency, we introduce a practical method to mitigate the challenges in AG-ReID posed by drastic viewpoint changes.
\item
\textbf{Novel framework.}
We present LATex, a novel feature learning framework that leverages attribute-based text knowledge with prompt-tuning strategies.
It not only reduces the resource requirement during training, but also extracts more discriminative features for AG-ReID.
\item
\textbf{Effective modules.}
We propose two effective modules, \emph{i.e.}, PACG and CPT.
PACG can effectively predict person attributes and generate attribute representations.
CPT integrates text knowledge by transforming attribute representations and view information into structured sentences.
\item
\textbf{Exhaustive validations.}
Extensive experiments on three AG-ReID benchmarks fully validate the effectiveness of our proposed methods.
\end{itemize}
\section{Related Works}
\subsection{View-Homogeneous ReID}
Person ReID is a long-standing task in computer vision and machine learning, drawing significant attention due to its wide range of real-world applications~\cite{zhang2024magic,hou2019vrstc,gu2020appearance,wang2025decoupled,wang2025mambapro}.
Previous research has primarily focused on view-homogeneous scenarios, where all cameras in the surveillance network are assumed to operate under the similar viewpoint.
Coarsely, the view-homogeneous ReID can be categorized into two types: ground-view and aerial-view.
In fact, the ground-view ReID has been widely researched with the support of various datasets, such as Market1501~\cite{zheng2015scalable}, MSMT17~\cite{wei2018person} and CUHK03~\cite{li2014deepreid}.
As a consequence, notable advancements have been achieved, primarily categorized into CNN-based methods and Transformer-based methods.
For CNN-based methods, Sun et al.~\cite{sun2018beyond} and Wang et al.~\cite{wang2018learning} enhance global feature representations by dividing the person image into several parts and extracting part-level features.
Furthermore, Luo et al.~\cite{luo2019bag} provide a strong ReID baseline by introducing some useful tricks.
Focusing on the computational efficiency, Quan et al.~\cite{quan2019auto} successfully construct a compact model, namely Auto-ReID, to obtain local discriminative features.
In recent years, many methods based on Vision Transformer (ViT) have emerged in the ReID community.
For example, He et al.~\cite{he2021transreid} first introduce Transformer into person ReID, achieving promising results.
Afterwards, many researchers further leverage Transformers to extract more discriminative person representations~\cite{zhu2023aaformer,zhang2021hat,yan2023learning,wang2024top,wang2025unity}.
Beside the spatial cues, Li et al.~\cite{li2024adaptive} extract high-frequency information of person images to obtain robust representations for ReID.
In terms of aerial-view ReID, UAV-Human~\cite{li2021uav} and PRAI-1581~\cite{zhang2020person} are the primary benchmarks.
As for advanced methods, Qiu et al.~\cite{qiu2024salient} introduce a key-point disentangling strategy for aerial-view ReID.
To address the challenge of person rotation, Wang et al.~\cite{wang2024rotation} propose a rotation exploration for aerial-view ReID.
However, these methods perform poorly under drastic viewpoint changes, which inevitably appear in AG-ReID.
\subsection{Aerial-Ground ReID}
Recently, advancements in Unmanned Aerial Vehicle (UAV) technologies have made it feasible to deploy dynamic cameras, enhancing surveillance coverage in regions with sparse ground camera networks.
However, it poses significant challenges due to the substantial viewpoint variations between UAV cameras and fixed ground cameras.
As a result, directly transferring previous view-homogeneous ReID methods often leads to suboptimal performance.
To address this issue, AG-ReID has been proposed as a new sub-task of person ReID.
To achieve the model training and evaluation, Nguyen et al.~\cite{nguyen2023ag} collect an outdoor scene dataset with person attribute annotations, namely AG-ReID.v1.
Afterwards, they extend the AG-ReID.v1 with more identities and viewpoints, as AG-ReID.v2~\cite{nguyen2024ag}.
Zhang et al.~\cite{zhang2024view} construct a large-scale synthesized AG-ReID benchmark, named CARGO.
Recently, Zhang et al.~\cite{zhang2024cross} consider video-based AG-ReID and contribute the first benchmark, named G2A-VReID.
Then, Nguyen et al.~\cite{nguyen2025ag} further contribute a large-scale video-based AG-ReID benchmark, named AG-VPReID.
As for technical methods, Nguyen et al.~\cite{nguyen2023ag,nguyen2024ag} propose multi-stream frameworks and use person attributes as auxiliary labels for supervision.
Based on ViT, Zhang et al.~\cite{zhang2024view} separate identity-related features from viewpoint-specific features by employing view tokens and an orthogonal loss.
Moreover, Wang et al.~\cite{wang2025secap} consider the person local features and introduce a prompt-based framework for better AG-ReID.
On the other hand, Wang et al.~\cite{dyu2024dynamic} employ a dynamic token selection strategy to focus on key person regions.
Although effective, these methods ignore the benefits of explicitly using person attributes for cross-view retrieval.
Based on the observations in Fig.~\ref{moti}, we notice that person attributes remain robust in complex scenarios, providing valuable information for discriminative features.
Motivated by this, we fully exploit this advantage to alleviate the viewpoint changes in AG-ReID tasks.
Technically, we propose a new feature learning framework named LATex that predicts and leverages attribute knowledge to achieve better performance with fewer trainable parameters.
\subsection{Prompt-Tuning in Person ReID}
Prompt-tuning aims to transfer the knowledge of pre-trained models to unseen domains via trainable prompts.
Moreover, it typically requires fewer computation resources than full fine-tuning, while also achieving superior performance.
This property makes prompt-tuning widely applicable across various tasks~\cite{jia2022visual,li2023clip,lester2021power}.
In the ReID field, Li et al.~\cite{li2023clip} first exploit vision-language models with prompt-tuning to address the lack of missing concrete text labels in image-based person Re-ID.
Yu et al.~\cite{yu2024tf} further propose text-free CLIP with prompt-tuning for video-based person ReID.
Wu et al.~\cite{wu2024enhancing} enhance visible-infrared person ReID with modality-aware and instance-aware visual prompt learning.
Wang et al.~\cite{liu2024distilling} introduce diverse prompt-tuning methods to distill CLIP for learning discriminative person shape representations.
Li et al.~\cite{li2024clip} propose person prompts and clothes prompts to learn cloth-agnostic features for cloth-changing person ReID.
Very recently, Wang et al.~\cite{wang2025secap} introduce self-calibrating and adaptive prompts for AG-ReID.
Yu et al.~\cite{yu2025climb} propose a hybrid CLIP-Mamba framework for person ReID.
Wang et al.~\cite{wang2025makes} adopt attribute prompt composition for object ReID.
Different from previous works, we not only employ learnable prompts as additional tokens to help pre-trained large models generalize to ReID domains, but also leverage prompt knowledge to further enhance discriminative features for AG-ReID.
\begin{figure*}[t]
    \centerline{\includegraphics[width=0.84\paperwidth]{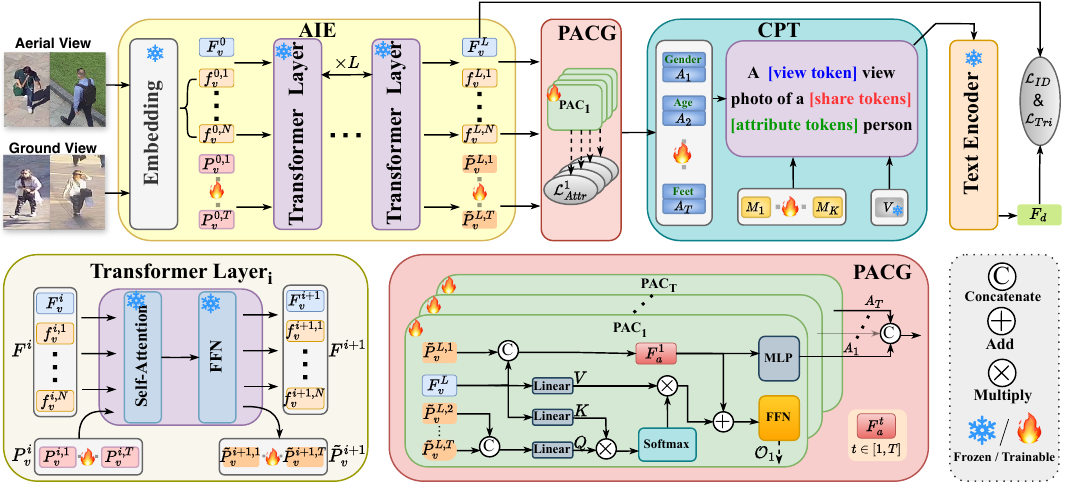}}
    \vspace{-2mm}
    \caption{The illustration of the proposed LATex framework.
    The Attribute-aware Image Encoder (AIE) first extracts global semantic features and attribute-aware features.
    Then, the Prompted Attribute Classifier Group (PACG) generates person attribute predictions and obtain specific representations of predicted attributes.
    Afterwards, the Coupled Prompt Template (CPT) transforms attribute representations and view information into structured sentences.
    Finally, the structured sentences are processed by the text encoder of CLIP to generate discriminative features for person ReID, integrated with global semantic features.}
    \vspace{-2mm}
    \label{overall}
\end{figure*}
\section{Proposed Method}
As shown in Fig. \ref{overall}, our LATex consists of three key components:
Attribute-aware Image Encoder (AIE), Prompted Attribute Classifier Group (PACG) and Coupled Prompt Template (CPT).
The details of them are as follows.
\subsection{Problem Definition}
We focus on the AG-ReID task where each person may be captured from different camera platforms, such as CCTV or UAV.
Our goal is to enable the model to correctly match the query image with the gallery image.
Formally, we define the problem as follows:
Given a training dataset \(\mathcal{C} = (\mathcal{C}^{Img},\mathcal{C}^{ID}, \mathcal{C}^{View} ) \), \(\mathcal{C}^{Img}\), \(\mathcal{C}^{ID}\) and \(\mathcal{C}^{View}\) denote the number of person images, identity labels and view labels, respectively.
We consider a model \( \mathcal{M} \) with trainable parameters \( \theta \) to extract discriminative representations from input images:
\begin{equation}
    F_i = \mathcal{M} (\mathcal{C}^{Img}_i,\mathcal{C}^{ID}_i, \mathcal{C}^{View}_i ; \theta),
\end{equation}
\begin{equation}
    F_j = \mathcal{M} (\mathcal{C}^{Img}_j,\mathcal{C}^{ID}_j, \mathcal{C}^{View}_j ; \theta),
\end{equation}
where \( F_i \) and \( F_j\) are the representations extracted by the model \( \mathcal{M} \), respectively.
Here, the input data consists of any two instances, \emph{i.e.}, \((\mathcal{C}^{Img}_i,\mathcal{C}^{ID}_i, \mathcal{C}^{View}_i)\) and \((\mathcal{C}^{Img}_j,\mathcal{C}^{ID}_j, \mathcal{C}^{View}_j )\), where \( i \neq j \).
Our training objective is to optimize the trainable parameters \( \theta \) such that during the inference phase, the following condition holds:
\begin{equation}
    \begin{cases}
        \mathcal{D}_{pos} = \mathcal{D} (F_i, F_j) & \text{if   } \mathcal{C}^{ID}_i = \mathcal{C}^{ID}_j, \\
        \mathcal{D}_{neg} = \mathcal{D} (F_i, F_j) & \text{if   } \mathcal{C}^{ID}_i \neq \mathcal{C}^{ID}_j, \\
        \mathcal{D}_{pos} \ll \mathcal{D}_{neg}, \\
    \end{cases}
\end{equation}
where \(\mathcal{D} (\cdot)\)  denotes a certain distance metric.
Given a query person image, we use this distance to match the gallery image corresponding to the same person identity.
\subsection{Attribute-aware Image Encoder}
To transfer the rich knowledge of CLIP to the ReID task and extract attribute information, we propose the Attribute-aware Image Encoder (AIE).
Previous Transformer-based approaches~\cite{nguyen2023aerial,zhang2024view} typically rely on full fine-tuning strategies, which lead to very high training costs.
To address this issue, we adopt prompt-tuning strategies to extract discriminative features with reduced training resource requirements.
Formally, given the input image \(\mathcal{V} \in \mathbb{R}^{H \times W \times 3}\) from different views, we embed \(\mathcal{V}\) to obtain the visual embedding \(F^0 = [F_{v}^{0}, f_{v}^{0}]\), where \(F_{v}^{0} \in \mathbb{R}^{C}\) is the class token and \(f_{v}^{0} \in \mathbb{R}^{N \times C}\) are patch tokens.
Here, \(C\) is the dimension of the token embeddings while \(N\) is the total number of patches.
Then, for the \(i\)-th Transformer layer \(\varOmega_{i}\), we denote $P^i_{v}\in\mathbb{R}^{\hat{T} \times C}$ as the learnable prompts, \emph{i.e.}, $P^i_{v} = \{P^{i,1}_{v},\cdots,P^{i,\hat{T}}_{v}\}$.
Here, the first $T$ prompts are treated as attribute-aware prompts.
The remaining $\hat{T}-T$ prompts are employed to support the fine-tuning of AIE.
The learnable prompts \(P^{i}_{v}\) are concatenated with \(F^{i}\), enabling the perception of attribute-specific information, and then passed through \(\varOmega_{i}\) as follows:
\begin{equation}
    [F^{i+1}, \tilde{P}^{i+1}_{v} ] = \varOmega_{i}([F^i, P^i_{v}]).
    \label{eq1}
\end{equation}
Here, [$\cdot$] means the concatenation operation along the token dimension.
Finally, we extract attribute prompts $ \tilde{P}^L_{v} \in \mathbb{R}^{T \times C}$ and the class token $F_{v}^{L}\in \mathbb{R}^{C}$ from the final layer of AIE for further processing.
\subsection{Prompted Attribute Classifier Group}
To explicitly predict person attributes using image information, we propose the Prompted Attribute Classifier Group (PACG), which integrates attribute information and exploits interdependencies among attributes.
More specifically, we define PAC\(_t\) as the classifier for the \( t\)-th attribute.
Then, the attribute feature \(F_a^t\) is defined as the concatenation of the global feature \(F_{v}^{L}\) and the corresponding attribute prompt \(\tilde{P}^{L,t}_{v}\):
\begin{equation}
F_{a}^{t} = [F_{v}^{L},\tilde{P}^{L,t}_{v}],
\label{eq3}
\end{equation}
where \(\tilde{P}^{L,t} \in \mathbb{R}^{C} \) is the \(t\)-th attribute prompt of \(\tilde{P}_{v}^{L}\).
To further enhance attribute-aware features, we utilize the interdependencies of different person attributes.
Formally, we denote other attribute prompts as $\hat{P}^{L}_{v} = \{ P_{v}^{L,1},\cdots,P_{v}^{L,t-1},P_{v}^{L,t+1},\cdots,P_{v}^{L,T} \}$.
Then, we can obtain interacted feature as follows:
\begin{equation}
    Q = W_q \hat{P}^{L}_{v}, K = W_k F_{v}^{L}, V = W_v F_{v}^{L},
\end{equation}
\begin{equation}
    \Theta (F_{v}^{L},\hat{P}^{L}_{v}) = \delta (\frac{QK^{T}}{\sqrt{C}})V,
\end{equation}
\noindent where $\Theta(\cdot)$ is the multi-head cross attention~\cite{vaswani2017attention}.
\(Q \in \mathbb{R} ^{C} \), \(K \in \mathbb{R} ^{C} \) and \(V \in \mathbb{R} ^{C} \) are generated by the corresponding projection matrix \( W_q \), \( W_k \) and \( W_v \), respectively.
\( \delta \) is the Softmax function.
The residual structure enables the model to process input features more flexibly~\cite{he2016Deep}.
As observed in daily life, certain person attributes exhibit strong correlations (\emph{e.g.}, gender and clothing), while others show weaker correlations (\emph{e.g.}, height and weight) or are nearly independent (\emph{e.g.}, ethnicity and gender).
Thus, we employ a residual-based Feed-Forward Network (FFN) to handle features obtained from two perspectives: \textbf{direct prediction} and \textbf{attribute dependency-based prediction}.
This design allows the FFN to adapt to diverse scenarios by effectively capturing attribute correlations, if they exist, while avoiding excessive noises and additional computational overhead caused by irrelevant attribute pairs:
\begin{equation}
    \mathcal{O}_{t}  = \Phi  (\Theta (F_{v}^{L},\hat{P}^{L}_{v}) + F_{a}^{t}),
\end{equation}
where $\Phi(\cdot)$ represents the FFN and \( \mathcal{O}_{t} \) is the final features used to predict attribute confidences.
In addition, to align the visual and textual representations in different feature spaces, we transform the visual embedding \(F_a^t\) and obtain attribute-based text representations as follows:
\begin{equation}
    A_{t} = \Psi(F_{a}^{t}),
    \label{eq4}
\end{equation}
where $\Psi$ is a Multi-Layer Perceptron (MLP).
The resulting representations serve as continuous textual tokens, which are fed into CLIP’s text encoder to enhance feature discrimination.
\subsection{Coupled Prompt Template}
Recently, large-scale vision-language models have delivered outstanding performance in many computer vision and natural language processing tasks.
As a fundamental component, text templates play an important role in text-based person ReID.
For example, Li et al.~\cite{li2023clip} utilize identity-specific prompt tokens to form a text template, \emph{i.e.}, ``A photo of a [learnable tokens] person.''
However, this kind of templates learn person attributes implicitly, lacking the explicit supervision and ignoring helpful view information.

To address the above issues, we propose a Coupled Prompt Template (CPT), which is presented as ``A [view token] view photo of a [shared tokens] [attribute tokens] person.''
This template not only couples identity-independent and attribute-aware information, but also leverages comprehensive knowledge of visual-language models.
More specifically, the view token $V$ is an instance-level text token, which depends on the view of person images.
For instance, the view token is designated as ``CCTV'' for images captured by ground views and ``UAV'' for those from an aerial view.
Functionally, the proposed framework is readily extensible to other viewpoints (e.g., a wearable device view) by simply introducing a new view token, requiring no architectural changes.
In addition, we formulate shared tokens as ``$[M_1, M_2, \cdots ,M_{K}]$'', where \(K\) is the total number of tokens.
These instance-shared tokens serve as register tokens~\cite{darcet2023vision} to enhance semantic feature representations.
As for attribute tokens, we denote them as ``$[A_1, A_2, \cdots ,A_{T}]$'', which are obtained by PACG and have rich attribute information.
By effectively utilizing these diverse tokens to form the structured sentence \(\mathcal{S}  \), our framework can fully leverage the useful information from person attributes.
Finally, we feed the sentence \(\mathcal{S}\) to the text encoder $ \mathcal{T} (\cdot) $ of CLIP to obtain the text feature $ F_{d} \in \mathbb{R}^C $:
\begin{equation}
    F_{d} = \mathcal{T} (\mathcal{S} ).
    \label{eq6}
\end{equation}
To improve the feature discrimination, we concatenate $ F_{d} $ and $ F_{v}^{L} $ for person retrieval.
With the CPT, our LATex can fully leverage the viewpoint invariance of person attributes to enhance the identity-related features.
\subsection{Loss Functions}
As illustrated in Fig. \ref{overall}, we employ multiple loss functions to optimize our framework.
For features obtained by AIE and CPT, we supervise them by the label smoothing cross-entropy loss~\cite{szegedy2016rethinking} and triplet loss~\cite{hermans2017defense}:
\begin{equation}
    \mathcal{L}_{ReID} = \lambda_{1}\mathcal{L}_{ID} + \lambda_{2}\mathcal{L}_{Tri}.
    \label{eq7}
\end{equation}
For attribute predictions, we employ the label smoothing cross-entropy loss to each \(\mathcal{O}_{t}\) obtained through PAC\(_t\):
\begin{equation}
    \mathcal{L}_{Attr}^{t} = - \frac{1}{|B|} \sum_{i=1}^{|B|}  c_i \log(\hat{c}_i).
    \label{eq8}
\end{equation}
Here, \(B\) is the batch size, $ c_i $ is the ground truth and $ \hat{c}_i $ denotes the corresponding attribute prediction.
Thus, the total loss for our framework can be given by:
\begin{equation}
    \scalebox{1.00}[0.95]{$ \mathcal{L} = \mathcal{L}_{ReID}^{AIE} + \mathcal{L}_{ReID}^{CPT} + \sum\limits_{t=1}^{T} \mathcal{L}_{Attr}^{t} $}.
    \label{eq9}
\end{equation}
\section{Experiment}
\subsection{Datasets and Evaluation Metrics}
\textbf{Datasets.}
We evaluate our methods on three AG-ReID benchmarks.
AG-ReID.v1~\cite{nguyen2023aerial} is a challenging dataset, consisting of 21,983 images captured by ground and aerial cameras of 388 identities, each annotated with 15 attributes.
Meanwhile, AG-ReID.v1 contains two protocols for evaluation, \emph{i.e.}, A→G and G→A.
AG-ReID.v2~\cite{nguyen2024ag} is an extended version of AG-ReID.v1, incorporating three views: aerial (A), ground (G), and wearable device (W).
Accordingly, the evaluations are expanded to include cross-view settings between A and W.
Specifically, AG-ReID.v2 consists of 100,502 images from 1,615 unique identities.
Finally, CARGO~\cite{zhang2024view} is a large-scale synthetic dataset, consisting of 108,563 images captured by five aerial cameras and eight ground cameras from 5,000 identities.
It is worth noting that there are no attribute annotations in this dataset.
For its protocols, CARGO contains four evaluation protocols, two of which are view-heterogeneous, while the other two are view-homogeneous.

\textbf{Metrics.} Following previous works, we employ the mean Average Precision (mAP)~\cite{zheng2015scalable} and Cumulative Matching Characteristic (CMC)~\cite{moon2001computational} at Rank-1 as evaluation metrics.
\subsection{Implementation Details}
Our proposed framework is implemented with PyTorch on one NVIDIA A100 GPU.
We use the pre-trained CLIP-Base-16 as our backbone.
All input images are resized to $256\times128$.
To enhance the generalization ability, we utilize multiple augmentation techniques such as random horizontal flipping, padding and random erasing~\cite{zhong2020random} for all inputs.
While the model is training, the mini-batch size is 128 consisted of 16 identities and 8 instances of each identity.
We fine-tune the model with Adam optimizer~\cite{kingma2014adam} with a base learning rate of 3.5e\(^{-4}\).
A learning rate scheduling strategy is employed, combining a warm-up phase with the cosine decay and a scaling factor of 0.01.
The total epoch is 120.
For the hyper-parameters, we set \( \lambda_{1} \) and \( \lambda_{2} \) in Eq. \ref{eq7} to 0.25 and 1.0, respectively.
The end-to-end training process takes 1.5 hours.

\textbf{LATex†.}
LATex† is a variant of LATex that adopts the full fine-tuning strategy to ensure a fair comparison with other full fine-tuning AG-ReID methods.
Specifically, we unfreeze all trainable parameters of the pre-trained CLIP's vision and text encoders and update them with a learning rate of 5e\(^{-6}\).
Other settings, including the learning rate decay strategy, optimizer, and the number of training epochs, are kept consistent with LATex.
The end-to-end training process takes about 2.5 hours.
\begin{table}[t]
    \centering
    \renewcommand{\arraystretch}{1.1}
    \caption{Performance comparison with different methods on AG-ReID.v1. The symbol † indicates results based on full fine-tuning strategies. The best and second-best results are highlighted in \textbf{bold} and \underline{underline}, respectively.}
    \vspace{-5mm}
    \begin{center}
    \resizebox{0.80\columnwidth}{!}{
    \begin{tabular}{c c c  c c}
        \toprule
        \multirow{2}{*}{Method} & \multicolumn{2}{c}{A→G} & \multicolumn{2}{c}{G→A}\\
        \cmidrule(r){2-3}\cmidrule(r){4-5}
        ~ & Rank-1 & mAP & Rank-1 & mAP \\
        \midrule
        OSNet\cite{zhou2021learning}             & 72.59 & 58.32 & 74.22 & 60.99\\
        BoT\cite{luo2019bag}                     & 70.01 & 55.47 & 71.20 & 58.83\\
        SBS\cite{he2023fastreid}                 & 73.54 & 59.77 & 73.70 & 62.27\\
        VV\cite{kumar2020strong}     & 77.22 & 67.23 & 79.73 & 69.83\\
        \midrule
        ViT\cite{dosovitskiy2020image}           & 81.28 & 72.38 & 82.64 & 73.35\\

        TransReID\cite{he2021transreid}          & 81.80 & 73.10 & 83.40 & 74.60\\

        PFD\cite{wang2022pose}                   & 82.30 & 73.60 & 82.50 & 73.90\\

        PHA\cite{zhang2023pha}                   & 79.30 & 71.30 & 81.10 & 72.10\\

        FusionReID\cite{wang2025unity}           & 80.40 & 71.40 & 82.40 & 74.20\\

        CLIP-ReID\cite{li2023clip}               & 79.44 & 70.55 & 84.20 & 73.05\\

        PCL-CLIP\cite{li2023prototypical}        & 82.16 & 73.11 & 86.90 & 76.28\\

        \midrule
        Explain\cite{nguyen2023aerial}           & 81.47 & 72.61 & 82.85 & 73.39\\

        VDT\cite{zhang2024view}                  & 82.91 & 74.44 & 86.59 & 78.57\\

        DTST\cite{dyu2024dynamic}                 & 83.48 & 74.51 & 84.72 & 76.05\\

        SeCap\cite{wang2025secap}                & 84.03 & \underline{76.16} & 87.01 & 78.34\\

        \midrule
        \rowcolor[gray]{0.9}
        \textbf{LATex}    & \underline{84.41} & 75.85 & \underline{88.88} & \underline{79.19}\\
        \rowcolor[gray]{0.9}
        \textbf{LATex†}    & \textbf{85.26} & \textbf{77.67} & \textbf{89.40} & \textbf{81.15}\\
        \bottomrule
    \end{tabular}
    }
    \vspace{-3mm}
    \end{center}
    \label{datav1}
\end{table}
\begin{table*}[h]
    \centering
    \renewcommand{\arraystretch}{1.2}
    \caption{Performance comparison on AG-ReID.v2. The superscript symbol † indicates results based on full fine-tuning strategies. The best and second-best results are highlighted in \textbf{bold} and \underline{underline}, respectively.}
    \resizebox{1.8\columnwidth}{!}{
    \begin{tabular}{c c  c c c c c c c}
      \toprule
      \multirow{2}{*}{Method}  & \multicolumn{2}{c}{A→C} & \multicolumn{2}{c}{A→W} & \multicolumn{2}{c}{C→A} & \multicolumn{2}{c}{W→A} \\
      \cmidrule(r){2-3} \cmidrule(r){4-5} \cmidrule(r){6-7} \cmidrule(r){8-9}
      ~                        & Rank-1 & mAP & Rank-1 & mAP & Rank-1 & mAP & Rank-1 & mAP \\
      \midrule
      Swin\cite{liu2021swin} & 68.76 & 57.66 & 68.49 & 56.15 & 68.80 & 57.70 & 64.40 & 53.90 \\
      HRNet-18\cite{wang2020deep} & 75.21 & 65.07 & 76.26 & 66.17 & 76.25 & 66.16 & 76.25 & 66.17 \\
      SwinV2\cite{liu2022swin} & 76.44 & 66.09 & 80.08 & 69.09 & 77.11 & 62.14 & 74.53 & 65.61 \\
      MGN(R50)\cite{2018Learning} & 82.09 & 70.17 & 88.14 & 78.66 & 84.21 & 72.41 & 84.06 & 73.73 \\
      BoT(R50)\cite{luo2019bag} & 80.73 & 71.49 & 86.06 & 75.98 & 79.46 & 69.67 & 82.69 & 72.41 \\
      BoT(R50)+Attributes & 81.43 & 72.19 & 86.66 & 76.68 & 80.15 & 70.37 & 83.29 & 73.11 \\
      SBS(R50)\cite{he2023fastreid} & 81.96 & 72.04 & 88.14 & 78.94 & 84.10 & 73.89 & 84.66 & 75.01 \\
      SBS(R50)+Attributes & 82.56 & 72.74 & 88.74 & 79.64 & 84.80 & 74.59 & 85.26 & 75.71 \\
      \midrule
      BoT(ViT)\cite{luo2019bag}    & 85.40 & 77.03 & 89.77 & 80.48 & 84.65 & 75.90 & 84.65 & 75.90 \\
      ViT\cite{dosovitskiy2020image} & 85.40 & 77.03 & 89.77 & 80.48 & 84.65 & 75.90 & 84.27 & 76.59 \\
      TransReID\cite{he2021transreid} & 88.00 & 81.40 & 90.40 & 84.50 & 87.60 & 80.10 & 87.70 & 81.10 \\
      FusionReID\cite{wang2025unity}  & 86.70 & 80.70 & 89.70 & 84.20 & 87.90 & 80.00 & 86.50 & 80.90 \\
      CLIP-ReID\cite{li2023clip}      & 85.36 & 79.79 & 89.14 & 84.23 & 85.64 & 79.08 & 86.50 & 79.55 \\
      PCL-CLIP\cite{li2023prototypical}  & 79.80 & 72.20 & 87.14 & 77.70 & 81.12 & 72.40 & 84.19 & 73.89 \\
      \midrule
      Explain\cite{nguyen2023ag}   & 87.70 & 79.00 & 93.67 & 83.14 & 87.35 & 78.24 & 87.73 & 79.08 \\
      VDT\cite{zhang2024view}      & 86.46 & 79.13 & 90.00 & 82.21 & 86.14 & 78.12 & 85.26 & 78.52 \\
      V2E(ViT)\cite{nguyen2024ag}  & \underline{88.77} & 80.72 & \textbf{93.62} & \underline{84.85} & 87.86 & 78.51 & \underline{88.61} & 80.11 \\
      SeCap\cite{wang2025secap}    &  88.12  & \underline{80.84}  & \underline{91.44} & 84.01 & \underline{88.24} & \underline{79.99} & 87.56 & 80.15 \\
      \midrule
      \rowcolor[gray]{0.9}
      \textbf{LATex} & 87.18 & 79.92 & 90.09 & 83.50 & 85.86 & 79.07 & 87.52 & \underline{80.93} \\
      \rowcolor[gray]{0.9}
      \textbf{LATex†} & \textbf{89.13} & \textbf{83.50} & 91.35 & \textbf{86.35} & \textbf{89.01} & \textbf{82.85} & \textbf{89.32} & \textbf{83.30} \\
      \bottomrule
    \end{tabular}
    }
    % \vspace{-1mm}
    \label{tab:datav2}
\end{table*}
\begin{table*}[h]
    \centering
    \renewcommand{\arraystretch}{1.2}
    \caption{Performance comparison on CARGO. The superscript symbol † indicates results based on full fine-tuning strategies.
    The best and second-best results are highlighted in \textbf{bold} and \underline{underline}, respectively.
    Performance is shown for view-heterogeneous protocols in \colorbox{red!10}{red}, and for view-homogeneous in \colorbox{blue!10}{blue}.}
    \begin{center}
    \resizebox{1.8\columnwidth}{!}{
    \begin{tabular}{c c c c c c c c c}
        \toprule
        \multirow{2}{*}{Method} & \multicolumn{2}{c}{\cellcolor{red!10}ALL} &  \multicolumn{2}{c}{\cellcolor{red!10}A\(\leftrightarrow\)G} & \multicolumn{2}{c}{\cellcolor{blue!10} G\(\leftrightarrow\)G} &  \multicolumn{2}{c}{\cellcolor{blue!10} A\(\leftrightarrow\)A}\\
        \cmidrule(r){2-3}\cmidrule(r){4-5} \cmidrule(r){6-7}\cmidrule(r){8-9}
         ~ & Rank-1 & mAP & Rank-1 & mAP & Rank-1 & mAP & Rank-1 & mAP\\
        \midrule
        SBS\cite{he2023fastreid} & 50.32 & 43.09 & 31.25 & 29.00 &72.31 & 62.99 & 67.50 & 49.73 \\

        PCB\cite{2021Learning} & 51.00 & 44.50 & 34.40 & 30.40 & 74.10 & 67.60 & 55.00 & 44.60 \\

        BoT\cite{luo2019bag}  & 54.81 & 46.49 & 36.25 & 32.56 & 77.68 & 66.47 & 65.00 & 49.79 \\

        MGN\cite{2018Learning} & 54.81 & 49.08 & 31.87 & 33.47 & 83.93 & 71.05 & 65.00 & 52.96 \\

        % VV\cite{2019Vehicle,kumar2020strong} & 45.83 & 38.84 & 31.25 & 29.00 & 72.31 & 62.99 & 67.50 & 49.73 \\
        VV\cite{kumar2020strong} & 45.83 & 38.84 & 31.25 & 29.00 & 72.31 & 62.99 & 67.50 & 49.73 \\

        AGW\cite{ye2021Deep} & 60.26 & 53.44 & 43.57 & 40.90 & 81.25 & 71.66 & 67.50 & 56.48 \\
        \midrule
        ViT\cite{dosovitskiy2020image} & 61.54 & 53.54 & 43.13 & 40.11 & 82.14 & 71.34 & 80.00 & 64.47 \\

        TransReID\cite{he2021transreid}  & 73.70 & 64.70 & 64.40 & 55.90 & 85.70 & 77.90 & \textbf{85.00} & \textbf{71.80} \\

        FusionReID\cite{wang2025unity}  & 67.90 & 61.50 & 48.30 & 53.10 & 85.70 & 79.40 & 80.00 & \underline{69.30} \\

        CLIP-ReID\cite{li2023clip} & 68.27 & \underline{64.25} & 55.62 & 53.83 & 84.82 & \textbf{80.80} & 75.00 & 65.42 \\

        PCL-CLIP\cite{li2023prototypical} & 67.31 & 60.93 & 54.37 & 51.43 & 84.82 & 76.00 & 70.00 & 60.75 \\

        \midrule
        VDT\cite{zhang2024view} & 64.10 & 55.20 & 48.12 & 42.76 & 82.14 & 71.59 & \underline{82.50} & 66.83 \\

        DTST\cite{dyu2024dynamic} & 64.42 & 55.73 & 50.63 & 43.39 & 78.57 & 72.40  & 80.00 & 63.31  \\

        SeCap\cite{wang2025secap} & \underline{68.59} & 60.19 &  \textbf{69.43} & \textbf{58.94} & \underline{86.61} & 75.42 & 80.00 & 68.08 \\

        \midrule
        \rowcolor[gray]{0.9}
        \textbf{LATex}            & 66.99 & 58.59 & 54.37 & 49.57 & 84.82 & 75.30 & 70.00 & 57.76 \\
        \rowcolor[gray]{0.9}
        \textbf{LATex†}            & \textbf{76.96} & \textbf{67.09} & \underline{66.87} & \underline{58.88} & \textbf{90.18} & \underline{79.90} & 80.00 & 69.06 \\
        \bottomrule
    \end{tabular}
    }
    \vspace{-4mm}
    \end{center}
    \label{datacargo}
\end{table*}
\begin{table}[t]
    \centering
    \renewcommand{\arraystretch}{1.3}
    \caption{Training and inference cost comparison of different methods.}
    \vspace{-1mm}
    \resizebox{.94\columnwidth}{!}{
    \begin{tabular}{c c c c c}
      \toprule
      Metric  & ViT & VDT & LATex & LATex†\\
      \midrule
      Trainable Params(M) & 86.24 & 85.90 & 35.97 & 122.00 \\
      GPU Memory(G)  & 0.075 & 0.078 & 0.079 & 0.108 \\
      Inference Speed(s/batch) & 0.35 & 0.41 & 0.67 & 0.67\\
      Flops(G)  & 11.37 & 11.46 & 15.33 & 15.33 \\
      \bottomrule
    \end{tabular}
    }
    \vspace{-3mm}
    \label{tab:cost}
\end{table}
\begin{table}[t]
    % \vspace{-1mm}
    \caption{Ablation results of key modules.}
    \vspace{-5mm}
    \begin{center}
    \resizebox{1.0\columnwidth}{!}{
    \renewcommand{\arraystretch}{1.3}
    \begin{tabular}{cc c c  c c  c c }
        \toprule
        \multirow{2}{*}{~} & \multicolumn{3}{c}{Module} & \multicolumn{2}{c}{A→G} & \multicolumn{2}{c}{G→A}\\
        \cmidrule(r){2-4} \cmidrule(r){5-6} \cmidrule(r){7-8}
                              & AIE          & PACG         & CPT          & Rank-1 & mAP & Rank-1 & mAP  \\
        \midrule
        A                     & $\checkmark$ & $\times$     & $\times$     & 81.69 & 72.36 & 83.89 & 74.78  \\

        B                     & $\checkmark$ & $\checkmark$ & $\times$     & 83.19 & 74.93 & 87.42 & 78.62 \\

        C                     & $\checkmark$ & $\checkmark$ & $\checkmark$ & \textbf{84.41} & \textbf{75.85} & \textbf{88.88} & \textbf{79.18} \\
        \bottomrule
    \end{tabular}
    }
    \vspace{-4mm}
    \end{center}
    \label{tab:abaModule}
\end{table}
\begin{table}[t]
    \caption{Performance comparison with different features.}
    \begin{center}
    \resizebox{.90\columnwidth}{!}{
    \renewcommand{\arraystretch}{1.3}
    \begin{tabular}{c c c c  c c}
        \toprule
        \multirow{2}{*}{~}  &\multirow{2}{*}{Feature} & \multicolumn{2}{c}{A→G} & \multicolumn{2}{c}{G→A} \\
        \cmidrule(r){3-4}\cmidrule(r){5-6}
                                & ~  & Rank-1 & mAP & Rank-1 & mAP    \\
        \midrule
        A                       & $ F_{v}^{L} $  & 83.19 & 74.93 & 87.42	& 78.62        \\

        B                      & $ [F_{v}^{L},F_a] $  & 83.85	& 75.07	& 88.05	& 78.74  \\

        C                      & $ [F_{v}^{L},F_{d}] $ & 84.41 & 75.85 & 88.88 & 79.18 \\

        D                   & $ F_{v}^{L} $  & 83.85	& 75.37	& 88.15	& 78.78        \\
        \bottomrule
    \end{tabular}
    }
    \vspace{-4mm}
    \end{center}
    \label{tab:abaAtrr}
\end{table}
\begin{table}[t]
    \centering
    \caption{Performance comparison with different backbones.}
    \renewcommand{\arraystretch}{1.2}
    \resizebox{0.9\columnwidth}{!}{
    \begin{tabular}{c c  c c c}
      \toprule
      \multirow{2}{*}{Method}  & \multicolumn{2}{c}{A→G} & \multicolumn{2}{c}{G→A} \\
      \cmidrule(r){2-3} \cmidrule(r){4-5}
      ~                        & Rank-1 & mAP & Rank-1 & mAP \\
      \midrule
      VDT(ViT-based) & 82.91 & 74.44 & 86.59 & 75.57 \\
      VDT(CLIP-based) & 78.78 & 68.40 & 77.44 & 68.68 \\
      \midrule
      LATex(ViT-based) & 74.93 & 62.75 & 73.18 & 63.84 \\
      LATex(CLIP-based) & 84.41 & 75.85 & 88.88 & 79.18 \\
      \bottomrule
    \end{tabular}
    }
    \vspace{-3mm}
    \label{tab:ababackbone}
\end{table}
\subsection{Performance Comparison}
We compare our proposed method with other person ReID methods on three AG-ReID benchmarks in Tab. \ref{datav1}, Tab. \ref{tab:datav2} and Tab. \ref{datacargo}.
Experiments on these AG-ReID benchmarks clearly show impressive performance of our proposed method.

On the AG-ReID.v1, our LATex achieves a Rank-1 accuracy of 88.88\% and an mAP of 79.19\% under the evaluation protocol G→A.
Compared with SeCap, our LATex presents improvements of 1.87\% in Rank-1 and and 0.85\% in mAP, respectively.
As for the evaluation protocol A→G, the performance of our LATex is 84.41\% Rank-1 and 75.85\% mAP, showing very competitive results.
The consistent improvements on two evaluation protocols clearly demonstrate the importance of leveraging attribute information in AG-ReID.
On two large-scale datasets, CARGO and AG-ReID.v2, our LATex achieves highly competitive performance.
It is worth noting that, there are no attribute annotations on CARGO.
To address the lack of attribute annotations, we remove PACG and directly use the output visual prompts from AIE as pseudo-attribute representations for CPT.
This adaptation ensures that our method can still leverage structural text prompts even in the absence of explicit attribute labels.
As a result, it can be used for attribute-missing benchmarks, such as CARGO.
The superior performance on CARGO fully validates its effectiveness in attribute-sparse domains.

To enable a fairer comparison with previous methods based on full fine-tuning, we introduce a LATex variant, namely LATex†.
Compared with LATex, LATex† achieves significant performance improvements across all benchmarks.
For example, on the CARGO dataset under all protocols, LATex surpasses DTST with a Rank-1 accuracy of 66.99\%, while LATex† further boosts this evaluation metric to 76.96\%, achieving an almost 10\% increase.
This fully demonstrates the scalability of our methods on powerful backbones.

\textbf{View-homogeneous ReID.}
As shown in Tab. \ref{datacargo}, CARGO provides protocols for person ReID under the same viewpoint, which we used to evaluate the performance of our LATex on view-homogeneous ReID tasks.
LATex outperforms DTST on the G\(\leftrightarrow\)G protocol.
LATex† achieves the superior overall performance among existing methods.
Notably, under the G\(\leftrightarrow\)G protocol, our LATex† is the first method to achieve a Rank-1 accuracy exceeding 90\%.
These results demonstrate that our method retains a strong generalization ability in view-homogeneous ReID tasks.

\textbf{CLIP-based ReID.}
Recently, CLIP has been used as a visual backbone for AG-ReID tasks.
To keep the advance, we compare our proposed method with some typical CLIP-based ReID methods in Tab. \ref{datav1}, Tab. \ref{tab:datav2} and Tab. \ref{datacargo}.
All models are trained with the full fine-tuning strategy.
We focus on this kind of comparisons on the CARGO benchmark, since CARGO does not comprise attribute annotations so that we can exclude the impact of additional information.
As shown in Tab. \ref{datacargo}, though these methods show impressive performance in view-homogeneous scenarios, they fail to handle the drastic viewpoint changes.
Consequently, they degrade significantly on CARGO.
In contrast, our LATex is able to address this issue, thus performs well in both view-homogeneous and view-heterogeneous settings.
\subsection{Training and Inference Cost Comparison}
In Tab. \ref{tab:cost}, we provide a comprehensive comparison of the training and inference costs on AG-ReID.v1.
Our proposed LATex significantly reduces trainable parameters compared with full fine-tuning methods like VDT and our LATex†.
Although LATex† achieves the highest performance, it requires substantially more trainable parameters and higher GPU memory usage.
In contrast, LATex offers a more efficient solution with competitive results.
Notably, the inference costs are determined by the model architecture.
Thus, the inference speed and FLOPs are the same between LATex and LATex†.
\begin{figure}[t]
    \centering
    \subfloat[A→G protocol]{\includegraphics[width=0.44\textwidth]{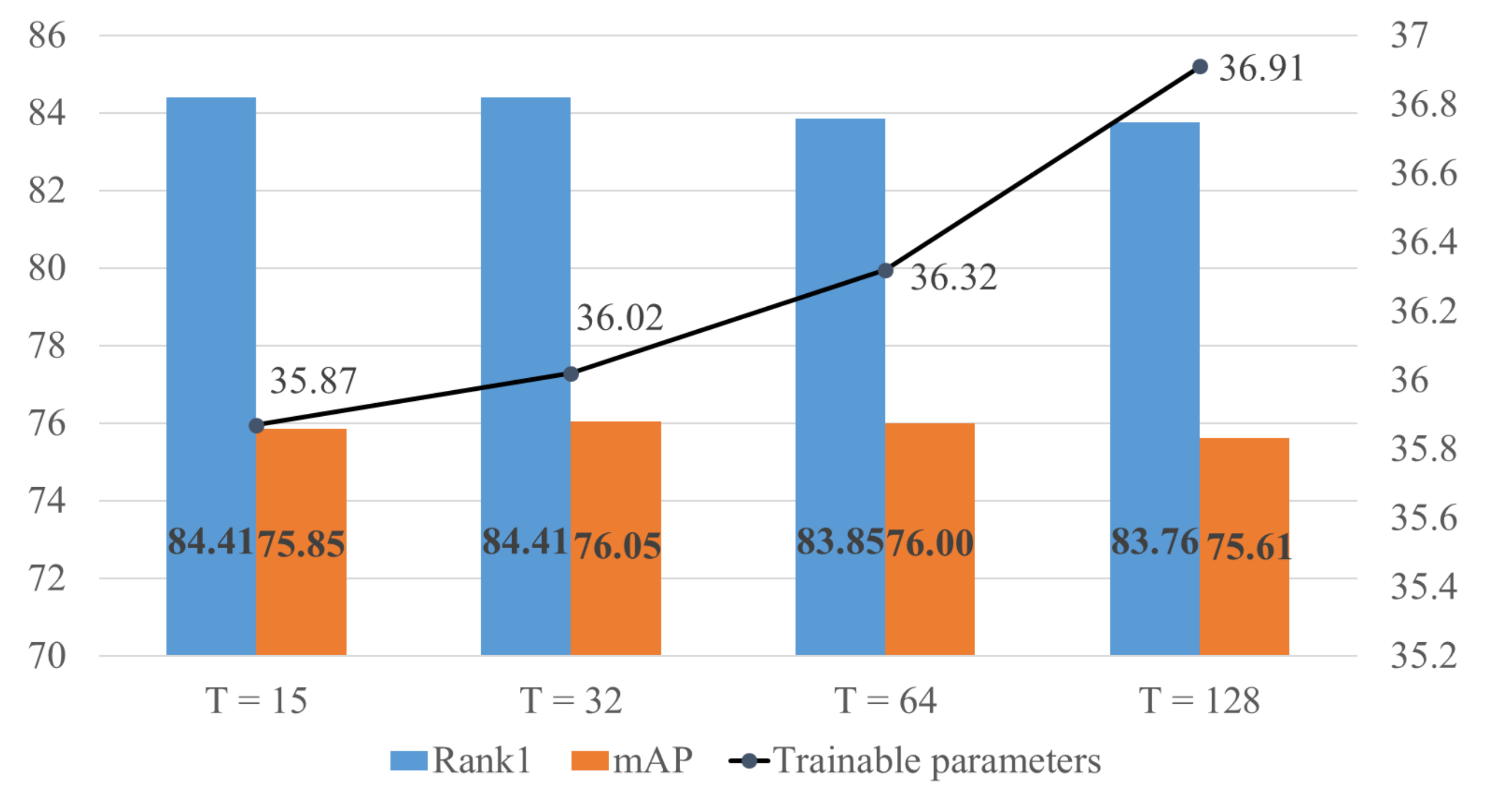}\label{fig:ptnum-a}}
    \hfill
    \vspace{-1mm}
    \subfloat[A→G protocol]{\includegraphics[width=0.44\textwidth]{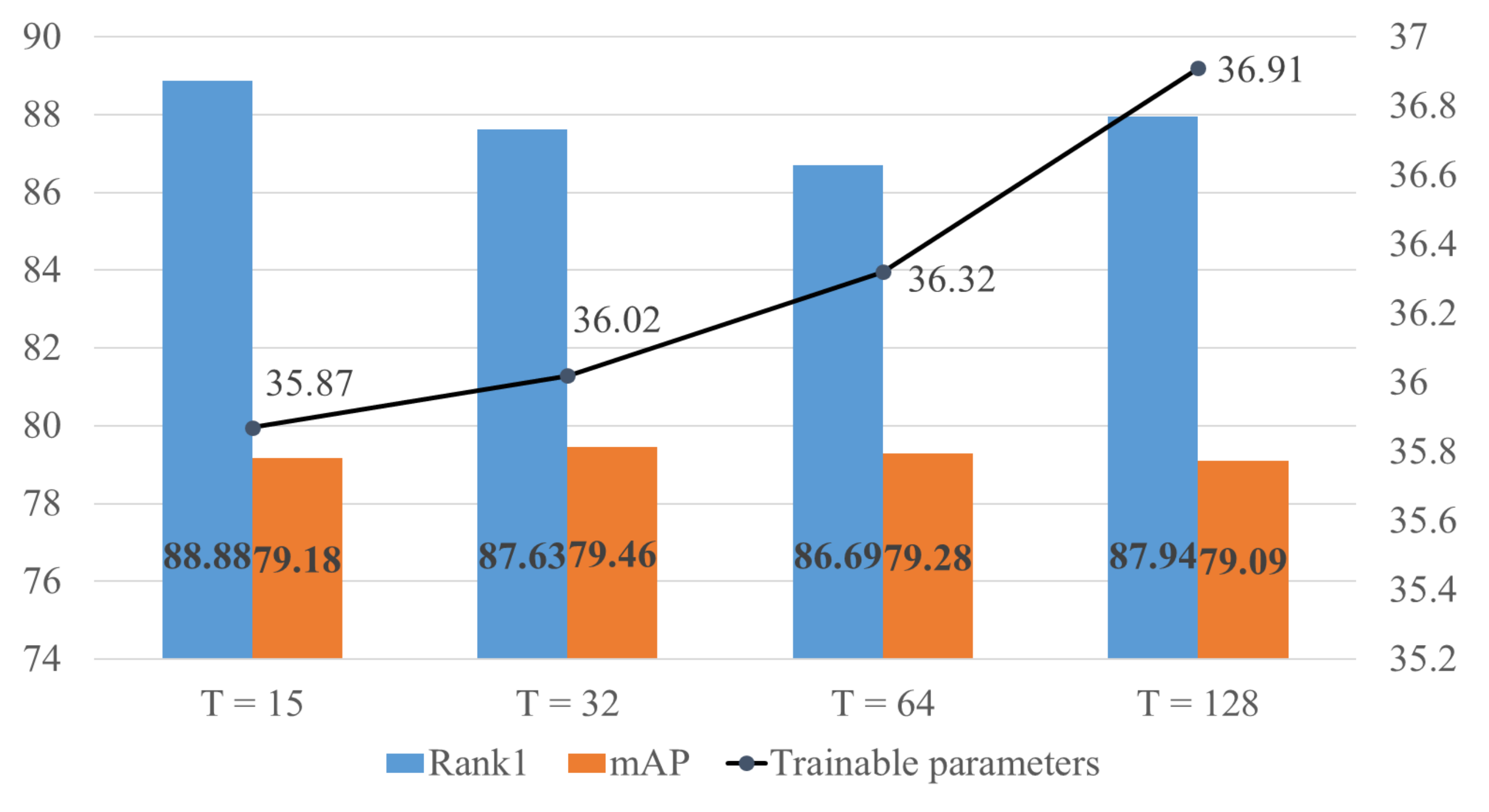}\label{fig:ptnum-b}}
    \caption{Performance with different numbers of prompts under two protocols.}
    \label{fig:ptnum}
\end{figure}
\begin{figure}[t]
    \vspace{-4mm}
    \centerline{\includegraphics[width = 0.42\paperwidth]{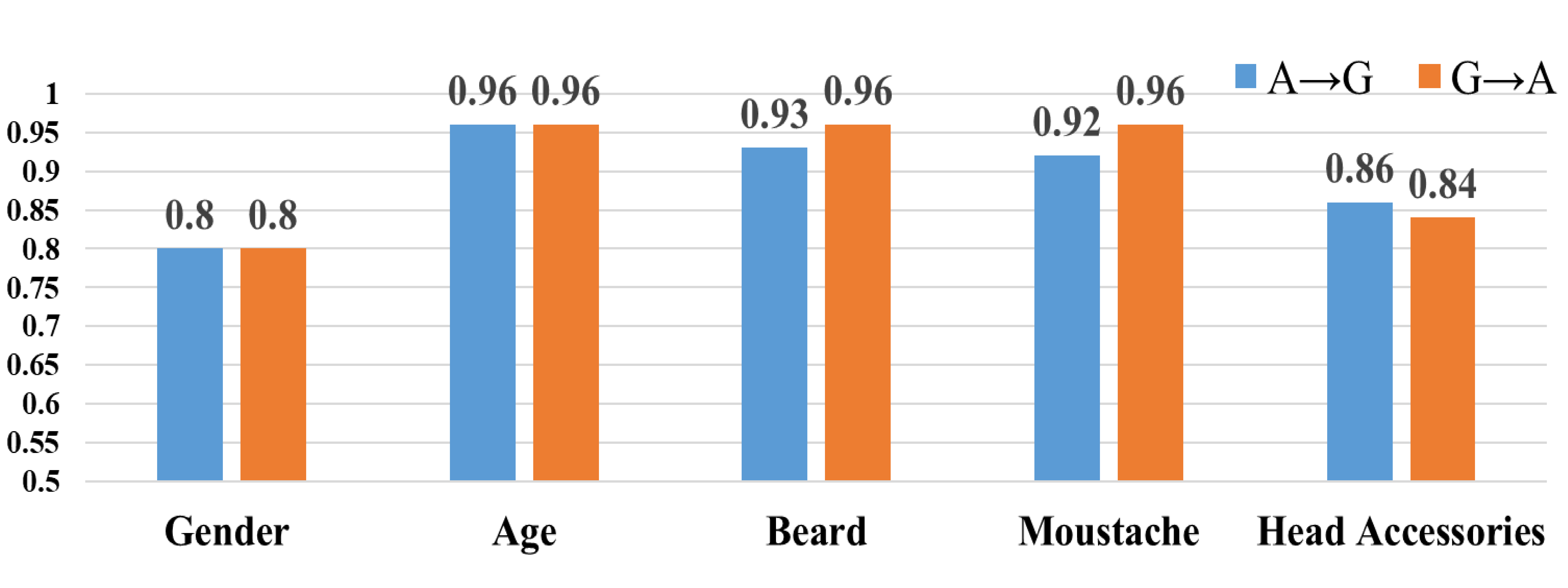}}
    \vspace{-4mm}
    \caption{Accuracy of attribute predictions in PACG.}
    \vspace{-3mm}
    \label{pred}
\end{figure}
\subsection{Ablation Studies}
To demonstrate the effectiveness of our proposed modules, we evaluate them on the AG-ReID.v1 dataset.
The results are shown in Tab. \ref{tab:abaModule}.
Furthermore, we conduct comprehensive and evaluations to investigate the details of our model design.

\textbf{Effect of Different Modules.}
In Tab. \ref{tab:abaModule}, Model A serves a our baseline.
It achieves a Rank-1 of 81.69\% and an mAP of 72.36\% under the protocol A→G, while obtaining a Rank-1 of 83.89\% and an mAP of 74.78\% under the protocol G→A.
With PACG, Model B increases the performance by 3.53\% Rank-1 and 3.84\% mAP under the protocol G→A.
Notably, its performance is already better than previous methods, such as VDT.
With CPT, Model C achieves the best result across all evaluation metrics.
These results clearly demonstrate the effectiveness of our key modules.

\textbf{Effect of Leveraging Person Attributes.}
We further validate the effectiveness of leveraging person attributes.
As shown in Tab. \ref{tab:abaAtrr}, Model A and Model B are implemented without the CPT module.
Model C and Model D are complete models.
The key difference lies in the features used for retrieval.
Specifically, \( F_a \) denotes the concatenation of all \( F_a^{t} \).
As can be observed, the performance of Model B is superior to that of Model A.
The performances of Model B and Model D are highly comparable.
The performance of Model C is the best.
These results clearly highlight the effectiveness of leveraging attribute-based text knowledge for AG-ReID.

\textbf{Effect of Backbones.}
We emphasize that the core idea of our LATex is to extract person attributes, embed them as pseudo-texts, and feed them into a text encoder to achieve robust person ReID.
With the strong vision-language alignment, we adopt CLIP as the backbone.
Considering that our approach uses a different backbone from previous methods, we evaluate the effect of various backbones and report the performance on AG-ReID.v1 in Tab. \ref{tab:ababackbone}.
As can be seen, CLIP-based VDT and ViT-based LATex show a significant performance drop.
The main reason is that they cannot fully leverage CLIP’s vision-language knowledge.
These results clearly demonstrate that our LATex's excellent performance stems from effectively utilizing CLIP's vision-language knowledge, rather than its inherent encoding capabilities.

\textbf{Effect of Trainable Prompts.}
Fig. \ref{fig:ptnum} shows the effect of the number of attribute prompts.
With the increase of trainable parameters, the performance of LATex remains stable, demonstrating its robustness to this factor.
\begin{table}[t]
    \centering
    \vspace{-0.5mm}
    \caption{Performance with different numbers of shared tokens.}
    \vspace{-0.5mm}
    \renewcommand{\arraystretch}{1.2}
    \resizebox{0.9\columnwidth}{!}{
    \begin{tabular}{c c  c c c}
      \toprule
      \multirow{2}{*}{Number}  & \multicolumn{2}{c}{A→G} & \multicolumn{2}{c}{G→A} \\
      \cmidrule(r){2-3} \cmidrule(r){4-5}
      ~ & Rank-1 & mAP & Rank-1 & mAP \\
      \midrule
      2 & 83.57 & 75.81 & 86.07 & 78.03 \\
      4 & 82.82 & 74.60 & 87.42 & 78.76 \\
      \textbf{8} & 84.41 & 75.85 & 88.88 & 79.19 \\
      12 & 83.66 & 75.00 & 85.03 & 78.29 \\
      16 & 84.51 & 75.28 & 87.11 & 78.78 \\
      \bottomrule
    \end{tabular}
    }
    \vspace{-1mm}
    \label{tab:abasharetoken}
\end{table}
\begin{table}[t]
    \centering
    \caption{Effectiveness of View Token. ``VT'' denotes the view token.}
    \renewcommand{\arraystretch}{1.2}
    \resizebox{0.90\columnwidth}{!}{
    \begin{tabular}{c c  c c c}
      \toprule
      \multirow{2}{*}{Method}  & \multicolumn{2}{c}{A→G} & \multicolumn{2}{c}{G→A} \\
      \cmidrule(r){2-3} \cmidrule(r){4-5}
      ~             & Rank-1 & mAP & Rank-1 & mAP \\
      \midrule
      VDT           & 82.91 & 74.44 & 86.59 & 78.57 \\
      LATex(w/o VT) & 83.94 & 75.13 & 87.11 & 78.30 \\
      LATex         & 84.41 & 75.85 & 88.88 & 79.19 \\
      GT Attributes & 98.78 & 98.37 & 100.00 & 99.36 \\
      \bottomrule
    \end{tabular}
    }
    \vspace{-1mm}
    \label{tab:abaviewtoken}
\end{table}

\textbf{Effect of Shared Tokens.}
Tab. \ref{tab:abasharetoken} analyzes the impact of the number of shared tokens.
We can observe that too few shared tokens would limit feature learning, while too many shared tokens may introduce noise.
Based on the results of ablation studies, we set the number of shared tokens to 8.

\textbf{Effect of View Tokens.}
Tab. \ref{tab:abaviewtoken} shows the effectiveness of view tokens.
Even though the variant without view token demonstrates competitive performance, explicitly incorporating view token s helps LATex further enhance its cross-view retrieval capabilities.
Since view tokens are camera-specific and can be pre-defined, we integrate these tokens into the CPT to achieve the optimal performance.

\textbf{Performance Upper Bound Analysis.}
To explore the performance upper bound, we remove PACG and directly incorporate attribute labels into the [attribute tokens] placeholder of CPT.
It simulates a scenario with perfect attribute predictions.
The last row of Tab. \ref{tab:abaviewtoken} shows that our method achieves notably high performance with ground truth attributes.
These results demonstrates a strong theoretical upper bound of our framework and its potential for further optimization.
\begin{figure*}[t]
    \centerline{\includegraphics[width=0.80\paperwidth]{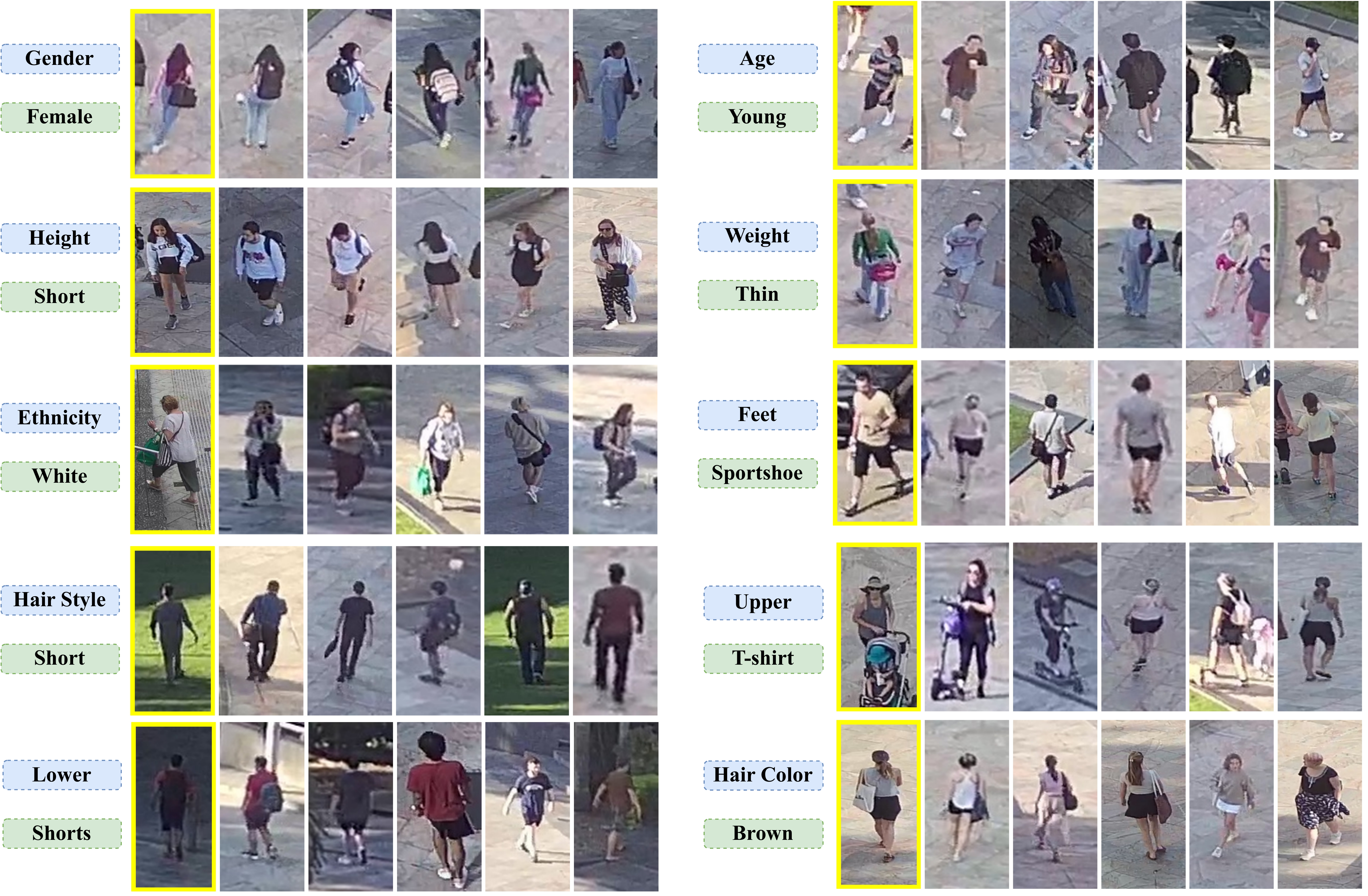}}
    \vspace{-2mm}
    \caption{The retrieval results using attribute features.
    Query images are marked with a yellow box.
    The corresponding attribute names and ground truths are displayed in blue and green boxes.}
    \vspace{-2mm}
    \label{fig:Attr_query}
\end{figure*}
\begin{figure}[t]
    \centerline{\includegraphics[width=0.4\paperwidth]{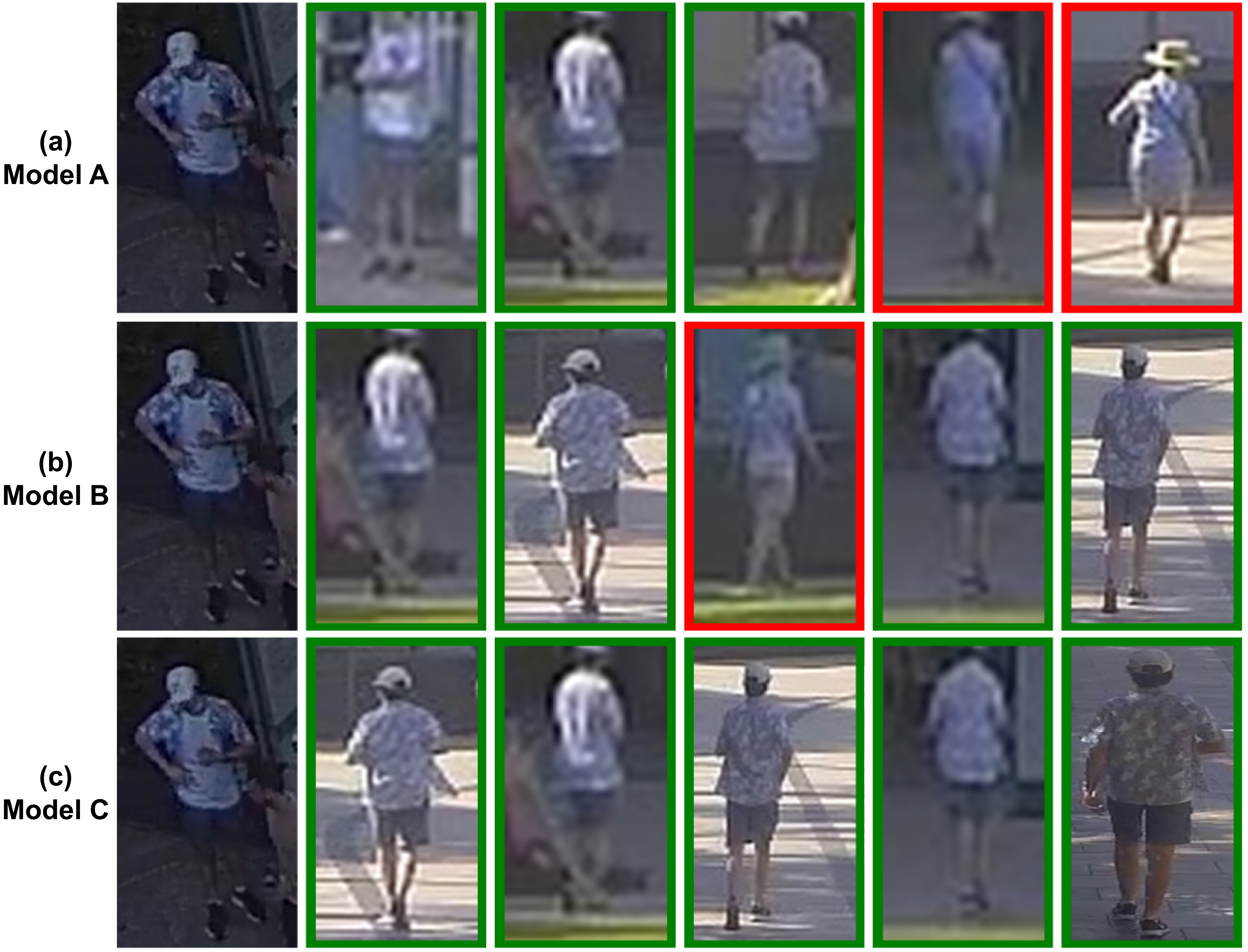}}
    \vspace{-2mm}
    \caption{Rank lists of different models defined in Tab. \ref{tab:abaModule}.
    Correctly retrieved images are marked with a green box, while incorrect ones with a red box.}
    \vspace{-2mm}
    \label{rank}
\end{figure}
\begin{figure}[t]
    \centerline{\includegraphics[width = 0.38\paperwidth]{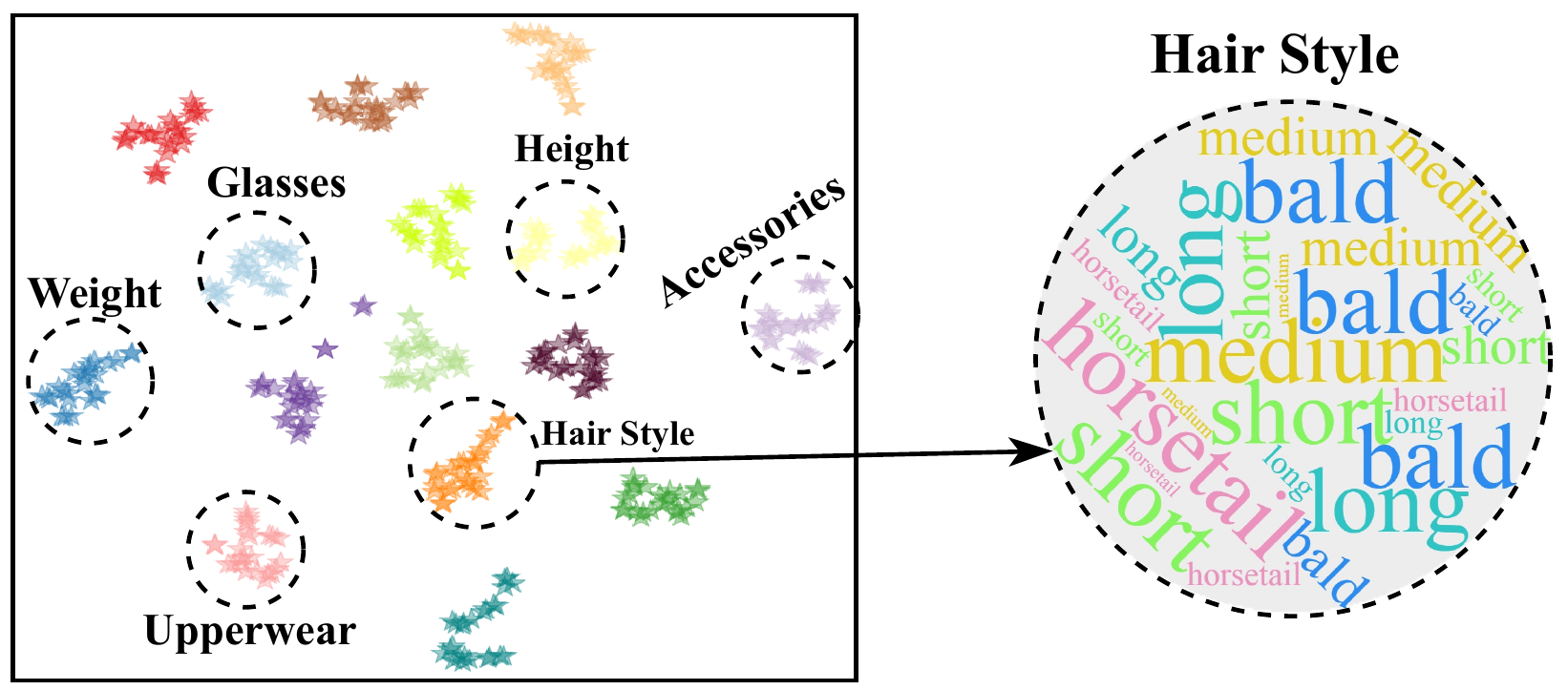}}
    \vspace{-2mm}
    \caption{Visualization of the attribute feature distributions with t-SNE\cite{van2008visualizing}.
    Different colors refer to different attributes, each comprising several fine-grained subcategories.}
    \vspace{-4mm}
    \label{fig:Attr_Tsne}
\end{figure}

\textbf{Accuracy Analysis of Attribute Predictions.}
Attribute predictions play an important role in our framework.
Fig. \ref{pred} shows the accuracy of some typical attributes predicted by PACG.
It can be observed that our PACG achieves outstanding performances in these person attributes.
These attributes provide discriminative information for person ReID.
\subsection{Visualization Analysis}
\textbf{Attribute Query Retrieval.}
Fig. \ref{fig:Attr_query} illustrates the retrieval results using the attribute features as the query.
Despite many challenges such as image blurriness and small key regions, LATex accurately retrieves persons sharing a specific attribute (\emph{e.g.}, upper clothes in our case).
These results show the exceptional capability of LATex in perceiving person attributes.

\textbf{Rank List Comparison.}
Fig. \ref{rank} compares the rank lists generated by different models, as defined in Tab. \ref{tab:abaModule}.
With the sequential addition of PACG and CPT, the rank lists become increasingly more accurate and discriminative.
This indicates that our model progressively acquires the ability to perceive person attributes, enabling it to better distinguish individuals with similar visual characteristics.

\textbf{Attribute Feature Distributions.}
Fig. \ref{fig:Attr_Tsne} illustrates the feature distributions of all attribute categories in the test set.
Each attribute category consists of several subcategories, which are finely distinguished by the corresponding PAC.
For example, the ``Hair Style" category includes various subtypes, such as ``Bald", ``Short", and ``Long''.
The results provide evidence that our method exhibits strong perception and discrimination capabilities for unseen samples across diverse attributes.
\section{Conclusions}
In this paper, we propose a novel feature learning framework, named LATex, for AG-ReID.
It adopts prompt-tuning strategies to integrate attribute-guided textual features from vision-language models.
To this end, we first propose an AIE to extract global semantic features and attribute-aware features.
Then, we propose a PACG to generate person attribute predictions and obtain representations of predicted attributes.
Finally, we design a CPT to transform attribute representations and view information into structured sentences for more discriminative features.
Extensive experiments on three AG-ReID benchmarks demonstrate the superior performance of our methods.
In the future work, we will improve person attribute prediction to further advance research in AG-ReID.

\bibliographystyle{IEEEtran}
\bibliography{reference}

\end{document}